%% file: egpaper_final.tex
\documentclass[10pt,twocolumn,letterpaper]{article}

\usepackage{cvpr}
\usepackage{times}
\usepackage{epsfig}
\usepackage{graphicx}
\usepackage{amsmath}
\usepackage{amssymb}
\usepackage{booktabs} 
\usepackage{latexsym}
\usepackage{mathtools}
\usepackage{url}
\usepackage{algorithm}
\usepackage{algorithmic}
\usepackage{enumerate}
\usepackage{multirow}
\usepackage{verbatim}
\usepackage{float}
\usepackage{subfig}
\usepackage{array}
\usepackage{tablefootnote}
\usepackage{bm}
\usepackage{wrapfig}
\usepackage{xcolor}
\usepackage{pbox}
\usepackage{footmisc}
\DeclareMathSizes{12}{30}{16}{12}

\newcolumntype{C}[1]{>{\centering\let\newline\\\arraybackslash\hspace{0pt}}m{#1}}

\usepackage[breaklinks=true,bookmarks=false]{hyperref}

\cvprfinalcopy 

\ifcvprfinal\fi
\begin{document}

\title{Visual Interest Prediction with Attentive Multi-Task Transfer Learning}

\author{Deepanway Ghosal\\
SUTD Singapore\\
{\tt\small deepanway\_ghosal@mymail.sutd.edu.sg}
\and
Maheshkumar H. Kolekar\\
Indian Institute of Technology Patna, India\\
{\tt\small mahesh@iitp.ac.in}
}

\maketitle

\begin{abstract}
   Visual interest \& affect prediction is a very interesting area of research in the area of computer vision. In this paper, we propose a transfer learning and attention mechanism based neural network model to predict visual interest \& affective dimensions in digital photos. Learning the multi-dimensional affects is addressed through a multi-task learning framework. With various experiments we show the effectiveness of the proposed approach. Evaluation of our model on the benchmark dataset shows large improvement over current state-of-the-art systems.
\end{abstract}

\section{Introduction \& Related Work}
In todays information age people create, share and consume millions of visual content everyday. With the increase of digital technology there has been the rise of huge photo and video collections in the Internet. Hence there is a need for creating effective image and video retrieval \& ranking systems from these rapidly growing digital repositories in order to perform content based recommendation, organization \& discovery of visual median collections \& databases. Traditionally the research on ranking visual contents has been performed based on aesthetics, quality and memorability \cite{dhar2011high,isola2013makes,ke2006design}. Although ranking \& retrieving contents based on the users interest would be a very relevant idea in this interactive era of Internet, not much studies has been performed on the area of visual interestingness prediction. Earliest work in visual interest prediction were primarily motivated by psychology \cite{psy1,psy2,psy3}. The idea that \textit{interest} is an \textit{emotion} was developed by Silvia et al. in \cite{psy4}. Research from this psychological perspective suggests that interestingness i.e. catching and holding human attention is related to uncertainty, complexity, conflict and novelty \cite{psy0}. The matter of interpersonal interestingness is also a very interesting topic in this regard. Work by authors in \cite{psy4} provides a detailed study about why different people perceives interestingness in different ways. Recently interestingness of image, gif and video content has been studied by authors in \cite{gygli2013interestingness,soleymani2015quest,grabner2013visual}.

In our current work, our objective is to determine the degree of various dimensions of visual interest in digital images. We propose a novel method involving image feature extraction using transfer learning, identifying the important \& relevant features using an attention based module and finally predicting the dimensions of interest using a multi-task learning based framework. Next we briefly discuss about transfer learning, attention based models and multi-task learning.

Transfer learning techniques are very frequently used in numerous machine learning problems. Convolutional neural networks which are at the core of transfer learning has been used in many interesting natural language processing \cite{ghosal2019dialoguegcn,akhtar2017multilayer,ghosal2017iitp}, computer vision \cite{bhatnagar2017classification} and audio signal processing \cite{ghosal2018music} problems. In this kind of systems, generally a deep convolutional net trained on the large scale (millions of data points) ImageNet data \cite{deng2009imagenet} is used. Although the original network is trained on ImageNet data, the weights learned by the network are meaningful because it is able to capture a wide variety of visual features which are then used for other visual recognition tasks. Weights of these pre-trained models (as they are previously trained on millions of images in ImageNet dataset) are then generally used to extract features for images in visual tasks where the amount training of training data is not so large. These features can then be used in various downstream machine learning tasks.  

Attention based models has also been quite popular in recent research. Although mostly used in natural language processing problems such as machine translation \cite{bahdanau2014neural,luong2015effective}, sentiment analysis \cite{ghosal2018contextual} \& document classification \cite{yang2016hierarchical}, they have also been applied to computer vision problems such as image captioning \cite{xu2015show}, image question answering \cite{yang2016hierarchical,lu2016hierarchical}, etc. In abstract terms, attentional models can be thought of models which are capable of focusing on part of a subset of the information they have been given as input \footnote{https://distill.pub/2016/augmented-rnns/}. 

Multi-task learning algorithms aims to leverage useful domain information contained in multiple related tasks using inductive transfer \cite{caruana1998multitask,akhtar2018multi,akhtar2019all,akhtar2019multi,ghosal2018deep}. The process of discovering or learning effective and useful features is key for any machine learning task. While neural networks trained using a single supervised objective often demonstrate very good performance, introducing multiple objectives for multiple related tasks, usually helps in improving the generalization performance of all the tasks. The representations shared across these multiple tasks helps in learning each of tasks in a better way. According to \cite{caruana1998multitask} multi-task learning encourages data augmentation, representation bias and provides parameter regularization. In the paradigm of deep neural networks, multi-task learning is typically done with either soft or hard parameter sharing of hidden layers. Recently some other promising models \cite{mtl1,mtl2,mtl3} have been proposed.


\section{Dataset \& Problem Setup}
\begin{figure*}[!ht]
	\begin{center} 
	   	\subfloat[\label{plot1}]
    	{
        	\includegraphics[width=0.23\textwidth, height=2.4cm]{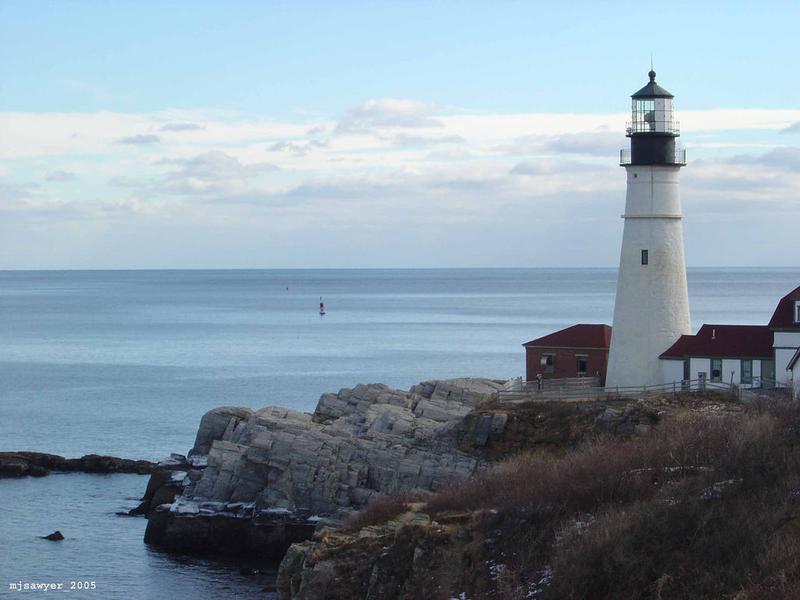}
    	}
	    \hspace{2px}
         \subfloat[\label{plot4}]
    	{
        	\includegraphics[width=0.23\textwidth, height=2.4cm]{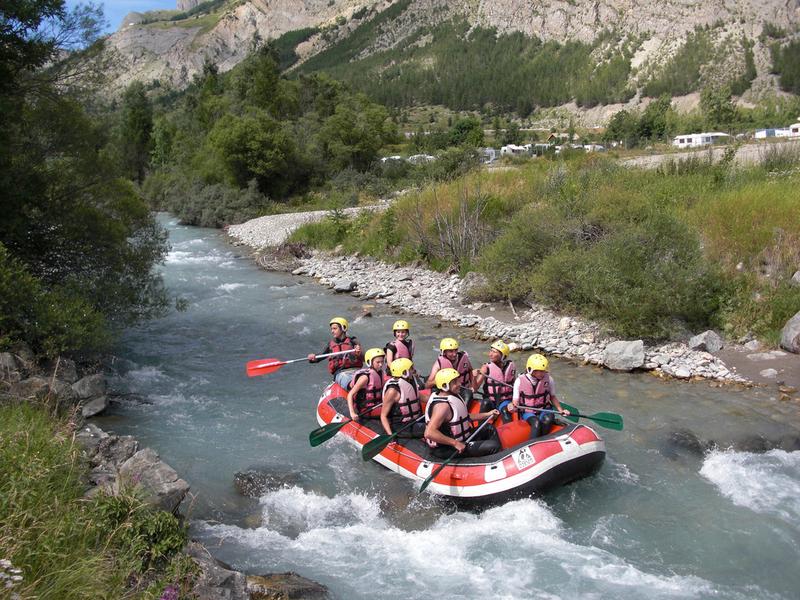}
    	}
	    \hspace{2px}
        \subfloat[\label{plot2}]
    	{
        	\includegraphics[width=0.23\textwidth, height=2.4cm]{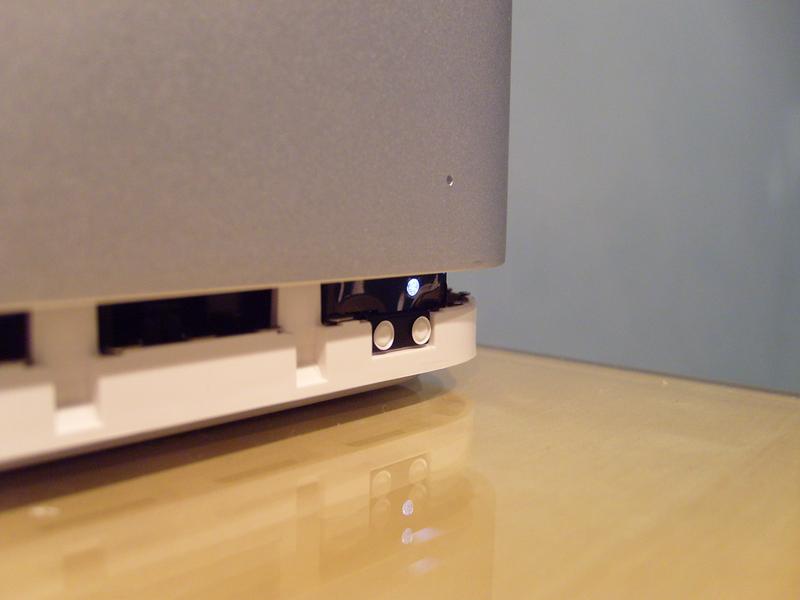}
    	}
	    \hspace{2px}
        \subfloat[\label{plot3}]
    	{
        	\includegraphics[width=0.23\textwidth, height=2.4cm]{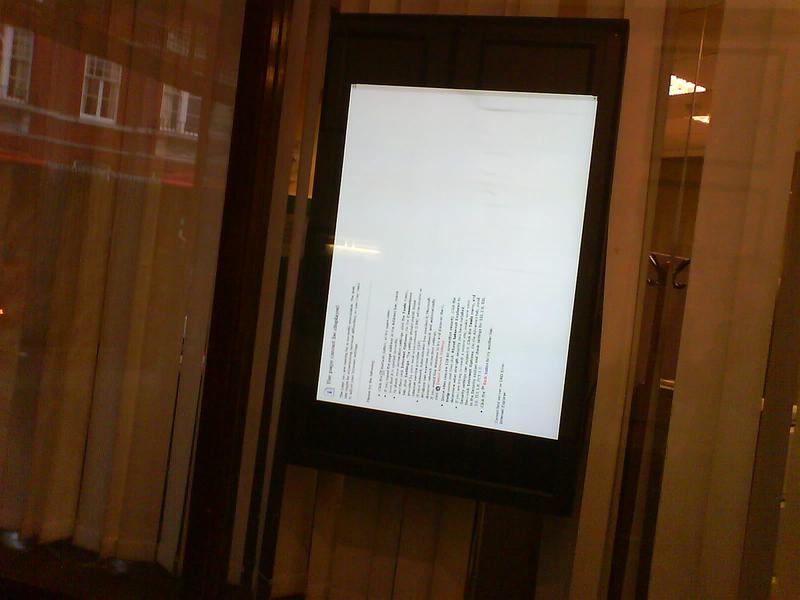}
    	}
	    \hspace{2px}
        
	\caption{Four example images from the dataset. (a) \& (b) has been annotated as very interesting whereas (c) \& (d) has been annotated as very less interesting.}
   	\label{fig-example}
	\end{center} 
\end{figure*}

We evaluate our model on the \textit{Visual Interest} dataset by Soleymani. \cite{soleymani2015quest}. The dataset contains 1005 digital photos from Flickr covering different topics, aesthetics and qualities. All the photos were collected using the Flickr API and were resized to $600 \times 400$ pixel dimension to match a 4:3 aspect ratio. A few example photos from the dataset are shown in Fig. \ref{fig-example}.

The photos were annotated via crowd-sourcing. A total of 66 annotators participated in the annotation process and each photo was annotated by 20 unique annotators. A set of 12 bipolar affective dimensions on a 7-point (1-7) differential scale were collected for each photo from each of the 20 unique annotators. This bipolar affective dimensions are as follows:

$\bullet$ i) Complexity (Complex - Simple)
$\bullet$ ii) Quality (Low quality - High quality)
$\bullet$ iii) Appeal (Appealing - Unappealing)
$\bullet$ iv) Naturalness (Natural - Staged)
$\bullet$ v) Pleasantness (Pleasant - Unpleasant)
$\bullet$ vi) Arousal (Arousing - Soothing)
$\bullet$ vii) Familiarity (Familiar - Unfamiliar)
$\bullet$ viii) Coping Potential (Easy - Hard (to understand))
$\bullet$ ix) Comprehensibility (Comprehensible - Incomprehensible)
$\bullet$ x) Coherence (Coherent - Incoherent)
$\bullet$ xi) Excitingness (Boring - Exciting)
$\bullet$ xii) General Interest (Interesting - Uninteresting) \newline
\indent In this work we aim to solve the problem of learning all the affective dimensions from the input image. It can be observed that quite a few affective dimensions are closely inter-related to each other. Hence we hypothesize that, a multi-task learning framework involving learning of all the affective dimensions would be able to achieve generalization by leveraging the inter-relatedness of multiple affective dimensions. We define the problem as joint learning of all the twelve affective dimensions from the input image. The intuition behind the introduction of multi-task learning is that, as the tasks are correlated then the joint-model can learn effectively from the shared representations.

\section{Methodology}
Our model has been designed as a generic framework for effectively modeling multi-dimensional affects in digital images. The architecture of the proposed \textit{attentive multi-task transfer learning} model can be sub-divided into four modules, Transfer Learning Feature Extraction (Section \ref{subsec1}), Primary Attention (Section \ref{subsec2}), Secondary Attention (Section \ref{subsec3}) and Multi-dimensional Affect Prediction (Section \ref{subsec4}).

\subsection{Transfer Learning Feature Extraction}
\label{subsec1}
First we divide the input image into $16$ $(4 \times 4)$ blocks of smaller sub-images. All the sub-images are equal dimensional ($150 \times 100$ pixels). We then extract features from each of the sixteen sub-images using four different pretrained ImageNet 
models. These pretrained models are 

\begin{itemize}
\item [i)] \textbf{Inception model}: We use the pretrained Inception-v3 model (on ILSVRC 2012 dataset \footnote{http://www.image-net.org/challenges/LSVRC/2012/}) introduced in \cite{szegedy2016rethinking} to extract a $2048$ dimensional feature vector (from the first non convolutional fully connected layer) for each sub-image. Compared to previous successful ImageNet models, the computational cost of the Inception-v3 model is much lower. The Inception-v3 model was designed in such a way so as to avoid representational bottlenecks early in the network while processing higher dimensional representations easily and keeping a certain balance between the width and height of the network.

\item [ii)] \textbf{Xception model}: The Xception model was proposed by authors in \cite{chollet2017xception} as an extension of the Inception-v3 model. The paper introduces the idea of depthwise separable convolution along with pointwise convolution. The authors show that, even though the capacity of the Xception model and the Inception-v3 model are same, the Xception model outperforms the Inception-v3 model in ImageNet dataset by using the model parameters more efficiently. We use the pre-trained Xception model as a feature extractor for the sub-images. A $2048$ dimensional feature vector is extracted for each sub-image.

\item [iii)] \textbf{Resnet model}: Deeper neural networks has more representational capability but are harder to train because of the vanishing/exploding gradient problem.  Authors in \cite{he2016deep} proposed a residual learning framework which makes the task of training deep neural networks significantly easier. In this Resnet model, the layers are reformulated to learn incremental or residual functions with reference to the input layers. The model adds more inter-layer connections in adjacent layers to boost the backward information flow during backpropagation. The Resnet model achieved state-of-the-art performances in ILSVRC \& COCO 2015 competitions. We use the pre-trained Resnet model as a feature extractor to extract $2048$ dimensional feature vectors for each of the sub-images.

\item [iv)] \textbf{Densenet model}: The Densenet model \cite{huang2017densely} extends the idea of Resnet model by introducing connections between each convolutional layer and every other convolutional layer in a feed-forward fashion. The Densenet model reduces the vanishing gradient problems, encourages feature reuse and reduces the number of train parameters substantially. The Densenet model produces excellent results in object recognition problems in a number of datasets. The pre-trained Densenet model is used as a feature extractor to extract $1920$ dimensional feature vectors for each of the sub-images. 
\end{itemize}

For the first three models the features extracted are $2048$ dimensional for each of the sixteen sub-images. For the fourth model we the extracted feature is $1920$ dimensional for each sub-image. Hence, we pad $128$ zeros at the end to make this feature 2048 dimensional. So, overall each image is represented by a $4 \times 16 \times 2048$ dimensional feature matrix ($4$ - the no. of pretrained models, $16$ - the no. sub-images, $2048$ -  the dimension of feature vector). Note that the spatial relation of the sub-images are not preserved in the feature matrix. 

\subsection{Primary Attention}
\label{subsec2}
Next an attention module is applied over the extracted transfer learning feature matrix. We call this the \textit{primary attention} module. An attention function is applied over the four pre-trained $16 \times 2048$ dimensional matrices. This attention module tries to find out which sub-images are actually important for the prediction of the different affective dimensions. The attention module learns to put more weight over the more contributing sub-images and less weight over the less contributing sub-images. So essentially the \textit{primary attention} module attends (puts required importance) over the 16 different sub-images. The operation is carried out as follows, 

\[
\left.
\begin{array}{lll}
&&u_{ij} = tanh(W(p) f_{ij} + b(p))\\ \\
&&\alpha_{ij} = \dfrac{exp(u_{ij}^T V(p))}{\sum_{}^{}\mathop{}_{\mkern-5mu j} exp(u_{ij}^T V(p))} \\ \\
&&f_i = \sum{}^{}\mathop{}_{\mkern-8mu j} \alpha_{ij} f_{ij} \\      
\end{array}
\right\}
\begin{array}{c}
\textit{for i = 1, 2, 3, 4 \&}\\
\textit{j = 1, 2, .., 16}\\
\end{array}
\]
\\
These operations are carried out for the four different pre-trained models (denoted by \textit{i = 1, 2, 3, 4}) and the sixteen different sub-image parts (denoted by \textit{j = 1, 2, .., 16}). In particular $f_{ij}$ represents feature extracted for the $j^{th}$ sub-image from $i^{th}$ pre-trained model. A single-layer MLP (multilayer perceptron) is applied on $f_{ij}$ to obtain its hidden representation $u_{ij}$. It is then used along with a sub-image level context vector $V(p)$ to obtain the attention weights $\alpha_{ij}$ for the sub-images. Those are then finally used to compute the attended sub-image feature $f_i$ for each pre-trained model (\textit{i = 1, 2, 3, 4} each denotes one pre-trained model). Along with $V(p)$, the weights \& bias of the MLP - $W(p)$ \& $b(p)$ are the learnable parameters in this \textit{primary attention} module. Note that, these \textit{primary attention} parameters ($V(p)$, $W(p)$ \& $b(p)$) are shared for all the pre-trained models. These parameters are learned through backpropagation during the training process.

\subsection{Secondary Attention}
\label{subsec3}
The \textit{secondary attention} module is applied over that output of the \textit{primary attention} module . The \textit{secondary attention} module attends over the four different pretrained models and learns to put weights over those according to their relative importance. This attention module consists of $12$ different attention operations, one each for each of the $12$ different affective dimensions. These attention operations are similar to that in the \textit{primary attention} module. Equations shown below illustrates the \textit{secondary attention} computation for a particular affective dimension $k$,

\[
\left.
\begin{array}{lll}
&&u_{i} = tanh(W(s)_k f_{i} + b(s)_k)\\ \\
&&\alpha_{i} = \dfrac{exp(u_{i}^T V(s)_k)}{\sum_{}^{}\mathop{}_{\mkern-5mu i} exp(u_{i}^T V(s)_k)} \\ \\
&&f = \sum{}^{}\mathop{}_{\mkern-8mu i} \alpha_{i} f_{i} \\
\end{array}
\right\}
\begin{array}{c}
\textit{for i = 1, 2, 3, 4}\\
\end{array}
\]

The attended feature vector $f$ is 2048 dimensional. For this particular affective dimension the learnable parameters are $W(s)_k$, $b(s)_k$ \& $V(s)_k$. As mentioned before, 12 different attention operations are carried out where each will have different $W(s)$, $b(s)$ \& $V(s)$. The final result is a set of 12 different 2048 dimensional vectors. 

\subsection{Multi-dimensional Affect Prediction}
\label{subsec4}
A set of $12$ different $2048$ dimensional vectors are obtained from the \textit{secondary attention} module. Note that the learned \textit{secondary attention} weights would be different for each of the $12$ different affective dimensions, as they will be learned separately. However the \textit{primary attention} weights would be same, as they will be learned in unison. Finally, each of the $2048$ dimensional vectors are passed separately through a few fully connected layers to produce the final affective score. Specifically, the fully connected layers consist of a hidden layer having $256$ neurons and then the output layer having a single neuron.

\subsection{Overall Model}

\begin{figure*}[!ht]
\begin{center}       
\includegraphics[width=1.0\textwidth, height=8cm]{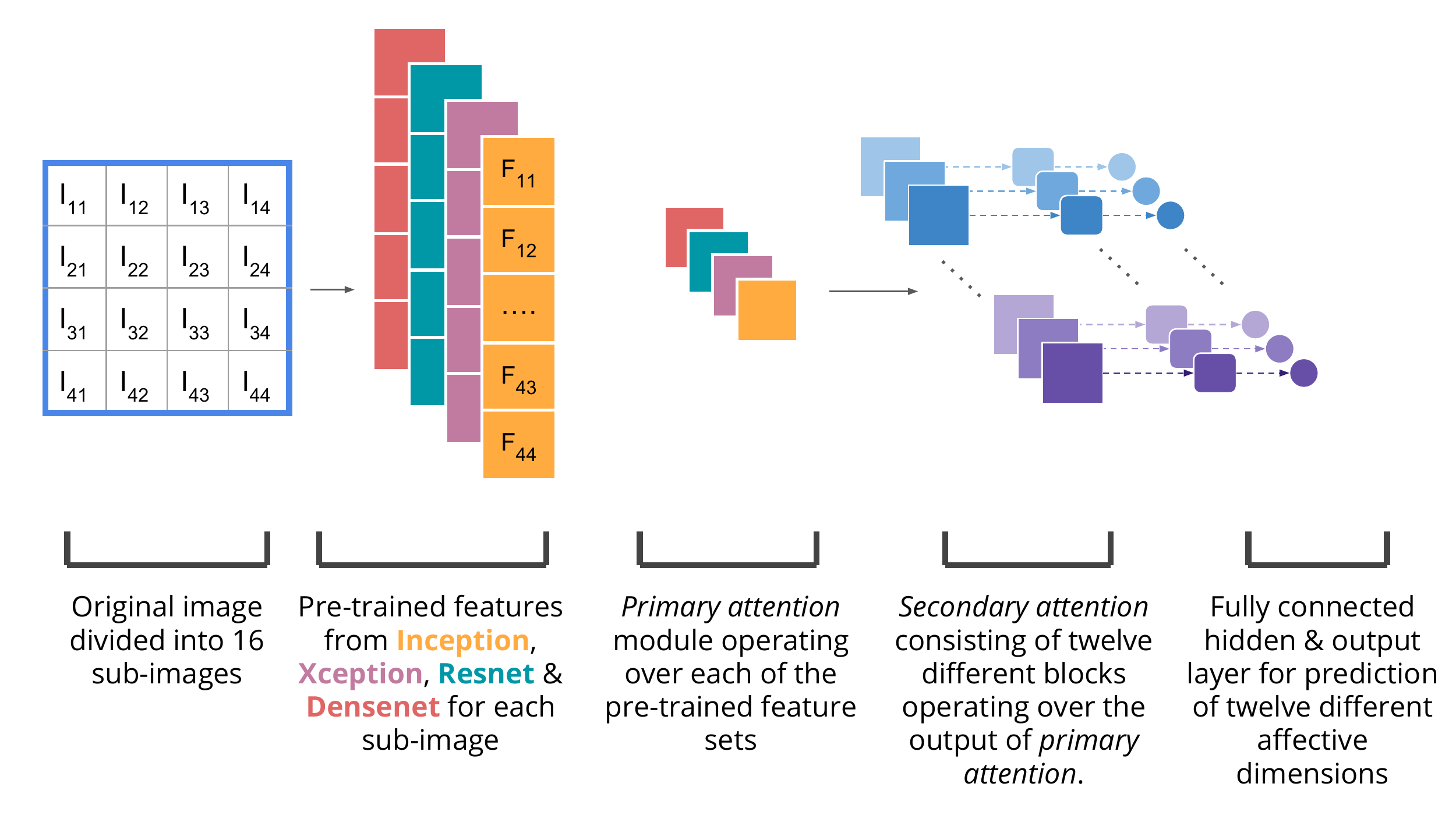}
\caption{The whole framework of the proposed \textit{Attentive Multi-task Transfer Learning} model.}
\label{network1}
\end{center} 
\end{figure*}

\label{subsec5}
An overall illustration of the proposed \textit{attentive multi-task transfer learning} model is shown in Fig \ref{network1}. It can be observed that the lowest level image features (transfer learning features from pretrained ImageNet models) and the \textit{primary attention} module is shared across the all the affective dimensions, whereas \textit{secondary attention} module operations are specific for each of the different affective dimensions. In this way, the model jointly tries to learn all the affective dimensions by sharing information in the lower layers of the neural network and finally having a task-specific layer which helps in learning the specific affective dimension. Naturally, this attentive multi-task model is trained jointly for all the output affective dimensions. 


\section{Experiments}
\subsection{Baseline Models}

In this section a few variants of the proposed \textit{attentive multi-task transfer learning} model are discussed. We call this models as baseline models. These baseline models largely consist of ablation case studies of the proposed framework.

\noindent $\bullet$ \textbf{Attentive Task-specific Transfer Learning Model:} 
As a baseline comparison we experimented with \textit{attentive task-specific transfer learning} models. The task-specific models are trained and tuned against one particular affective dimension at a time. Hence we have 12 different task-specific models, one for each affective dimension. Each of the model consists of the pretrained ImageNet features for the sub-images, the first attention block attending over the sub-images, the next attention block attending over the pretrained models, the fully connected hidden layer and finally the output layer denoting the specific output attribute. 

\noindent $\bullet$ \textbf{Non-attentive Multi-task Transfer Learning Model:}
A non-attentive model would require the input to be a 1D feature vector and subsequent layers to be fully connected in nature. A good idea is to turn the original input feature matrix of $4\times 16 \times 2048$ dimension (in the proposed attentive model) to a 1D feature vector and giving that as input to this non-attentive model. But this would require the feature vector to have a dimension of $131072$ which is quite large and computationally quite expensive in practice, when taken into account that the subsequent layers would be fully connected layers. For example if the first hidden layer has 512 nodes, then there are approximately 67 million trainable parameters between the input layer and the first hidden layer along with more parameters in the next hidden layers.

So for this model in particular, instead of extracting features from sub-images, we extract features from the whole image. Feature extraction from all the four pre-trained models are agnostic to the image shape (because the primary layers are convolutional in nature \& global max-pooling is finally used to have a constant feature size). Hence we extracted $2048$, $2048$, $2048$ \& $1920$ dimensional features from the four pre-trained models and concatenated them to have a final feature vector of dimension $8064$. This is the input for our \textit{non-attentive multi-task transfer learning} model. A series of fully connected operations are then performed to obtain the multiple affective dimension output. In particular we use, two hidden layers which are shared across all the affective dimensions and then two hidden layers \& the output layer which is specific for each affective dimension.

\noindent $\bullet$ \textbf{Attentive Multi-task Non-transfer Learning Model:}
We created four convolutional neural networks resembling the four Imagenet models (Inception, Xception, Densenet, Resnet) and initialized the weights of the networks randomly. This models are then used to extract features from the sub-images. Then \textit{primary attention}, \textit{secondary attention} and \textit{multi-dimensional affect prediction} is carried out in a way similar to the original model (viz. Section \ref{subsec2}, Section \ref{subsec3} \& Section \ref{subsec4}).

Note that the initial weights of the resembling neural networks are random, thus we need to train these as well along with the weights in the attention layers \& the prediction layer. Hence the whole pipeline (from feature extraction using the convolutional model to multi-dimensional affect prediction) has to be trained jointly. This is the contrasting difference between the non-transfer learning model here and the original transfer learning model. In the original transfer learning model, the pre-trained models are only used to extract features from the sub-images. The weights of those pre-trained are already capable of meaningful feature extraction, hence they are not needed to be trained further. Once feature extraction is done, only weights in the attention layers \& the prediction layer are trained using backpropagation.

\noindent $\bullet$ \textbf{Non-attentive Task-specific Non-transfer Learning Model:} 
This is the model without any of attention modules, multi-task learning or transfer learning. For this model we created four convolutional neural networks resembling the four Imagenet models (Inception, Xception, Densenet, Resnet) with randomly initialized weights. The models are used to extract $2048$, $2048$, $2048$ \& $1920$ dimensional features from the whole image. These four feature vectors are then concatenated to obtain a $8064$ dimensional feature vector. Next a series of fully connected operations are performed to obtain the output affective dimension. As this is task-specific, we trained twelve different \textit{non-attentive task-specific non-transfer learning} models, one for each affective dimension.

\subsection{Training Details}
We took \textit{mean} of the labels provided by the 20 annotators and scaled them between [-0.5, +0.5] following \cite{soleymani2015quest}. For further analysis we also report results for models when trained against \textit{median} of the labels scaled between [-0.5, +0.5]. For evaluation we compute the \textit{r-squared} ($r^2$), \textit{pearson correlation} ($\rho$) and \textit{rmse} (root mean squared error) metrics. For all our models we use \textit{dropout} \cite{srivastava2014dropout} of 0.5 in the fully connected layer and \textit{tanh} activation in the output layer. The models are trained with \textit{mean squared error} loss function with \textit{Adam (learning rate = 0.001)} optimizer \cite{kingma2014adam}. Note that in the multi-task learning framework the model is trained against all the twelve affective dimensions. Hence the overall \textit{mean squared error} loss value would be the sum of the twelve individual \textit{mean squared error} loss values. We train our model for 60 epochs with batch size of 32 with early stopping (patience=10) \cite{caruana2001overfitting}. Due to randomness phenomenon in neural network training, we train each model five different times and report the average results.

\section{Results \& Analysis}
\input{main_results}
First, we report our five-fold cross validation results for the \textit{attentive multi-task transfer learning} model in Table \ref{result-multi-mean}. The results in the left hand side of the table are for the model when trained against the scaled \textit{mean} labels, whereas results in the right hand side of the table are for the model when trained against the scaled \textit{median} labels. We found that scaled mean labels mostly produces better results compared to scaled median labels. 

As models trained against mean labels produces better results in practice, we report results for baseline models when trained against the mean labels with five-fold cross validation. Table \ref{result-abla1} shows results for the \textit{attentive task-specific transfer learning} models \& \textit{non-attentive multi-task transfer learning} model. Table \ref{result-abla2} shows results for the \textit{attentive multi-task non-transfer learning} model \& \textit{non-attentive task-specific non-transfer learning} models.

\subsection{Comparison with Baseline Models}
We observe that the proposed \textit{attentive multi-task transfer learning} model outperforms the \textit{attentive task-specific transfer learning} models. The multi-task learning setup consists of a single joint model which learns over all the twelve affective dimensions, whereas the task specific learning setup consists of twelve different models which learns over each of the affective dimensions separately. Even though the multi-task learning setup consists of a single model, we observe that it outperforms the task specific learning models in almost all the cases (and with significant margins for multiple affective dimensions).

Similar is the case for \textit{attentive multi-task transfer learning} model and \textit{non attentive multi-task transfer learning} model. Both are multi-task learning models, but removing the attention module causes a significant drop in performance. Thus it can be said that the attention module is an integral part of the proposed framework. By jointly attending over the sub-images and the pre-trained models it is capable of capturing visual cues which are important for interest prediction.

Likewise there is a very significant performance drop in the \textit{attentive multi-task non-transfer learning} model compared to the \textit{attentive multi-task transfer learning} model. Performance of this particular model clearly indicates that the features extracted from the pre-trained models are probably the most important component in the whole framework.

These baseline models also constitutes of the ablation study of the proposed \textit{attentive multi-task transfer learning} model. By eliminating one major component at a time we show the significance of each of the components - i) \textit{transfer learning feature attraction} ii) \textit{attention module} \& iii) \textit{multi-dimensional affect prediction}. Although each of these are very important in the final prediction, it can be said from the results in Table \ref{result-abla1} \& \ref{result-abla2} that \textit{transfer learning feature extraction} is the most contributing component in the whole framework.

\subsection{Comparison with State-of-the-art Systems}
\input{comparative_results}
We compare the performance of our proposed \textit{attentive multi-task transfer learning} model in the \textit{Visual Interest} dataset with \cite{soleymani2015quest}. The results are shown in Table \ref{result-compare}. Authors in \cite{soleymani2015quest} reported results only for eight of the twelve affective dimensions (viz. Table \ref{result-compare}). The authors first extracted various low-level visual features from the images and then trained a regression algorithm with sparse approximation to predict the different attributes. The extracted feature set include: max-pooled histogram of oriented gradients (HOG), max-pooled local binary patterns (LBP), color name feature, bag of word representation of GIST descriptors \cite{khosla2012memorability}, color histogram, contrast, naive arousal score \cite{machajdik2010affective}, edge distribution \cite{gygli2013interestingness}, jpeg compression rate from an uncompressed image and spatial pyramids of sift histograms \cite{lazebnik2006beyond}.

Our model achieves better performance than Soleymani. \cite{soleymani2015quest} for all the affective dimensions across all the performance metrics. The improvement in performance in terms of \textit{r-squared} \& \textit{pearson correlation} metric is considerably large for all the affective dimensions.

\subsection{Analysis of Attention Mechanism and Architecture}
\begin{figure*}[!ht]
	\begin{center} 
	   	\subfloat[\label{plot1}]
    	{
        	\includegraphics[width=0.23\textwidth, height=2.4cm]{image2.jpg}
    	}
	    \hspace{2px}
         \subfloat[\label{plot4}]
    	{
        	\includegraphics[width=0.23\textwidth, height=2.4cm]{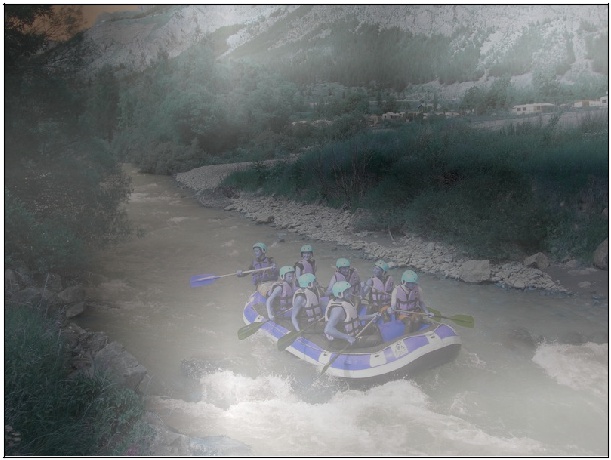}
    	}
	    \hspace{2px}
        \subfloat[\label{plot2}]
    	{
        	\includegraphics[width=0.23\textwidth, height=2.4cm]{image4.jpg}
    	}
	    \hspace{2px}
        \subfloat[\label{plot3}]
    	{
        	\includegraphics[width=0.23\textwidth, height=2.4cm]{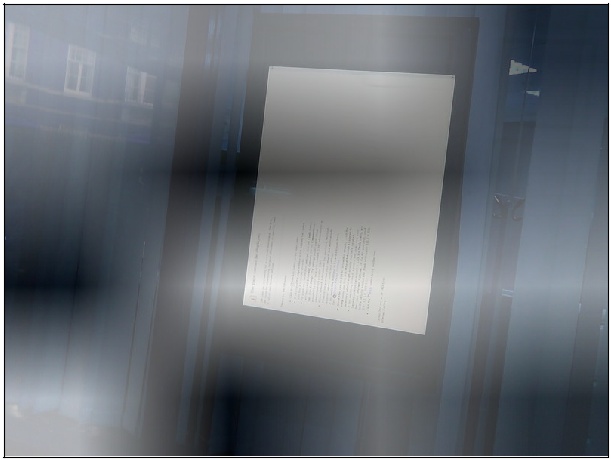}
    	}
	    \hspace{2px}
        
	\caption{(a) \& (c) Two images in the dataset whose annotated affective dimensions are very interesting \& very less interesting respectively. The proposed model predicted all the affective dimensions with near precision for these two images. (b) \& (d) \textit{Primary attention} weights when the images are in the validation split during cross-validation. Darker parts signify less attention weights whereas lighter parts signify more attention weights. In (b), the interestingness is driven by the more attention weights in the areas of the river, mountain \& the people rafting and less weights in the area of shrubs. In (d), the very less interestingness is driven by more or less random weights all over the screen.}
   	\label{fig-attention}
	\end{center} 
\end{figure*}

We further analyze the effect of attention module for affect prediction. Fig \ref{fig-attention} illustrates two such example cases. It can be seen that model has learned to put required attention weights in the sub-images for successful prediction of the different affective dimensions.

This particular choice of architecture (\textit{primary attention} over the sub-images \& \textit{secondary attention} over the pre-trained models) in the proposed model is inspired by our hypothesis that prediction of the all the different affective dimensions will be more or less affected by the same areas of the image. The \textit{primary attention} module essentially learns to find the sub-parts of the image which will be more important for the prediction of the affective dimensions. Rather than same \textit{primary attention} parameters ($V(p)$, $W(p)$ \& $b(p)$) for all the pre-trained models, one might hypothesize that separate primary attention parameters (one set of $V(p)$, $W(p)$ \& $b(p)$ for each pre-trained model) would be more appropriate. This may seem a more flexible model in terms of attention computation, but we didn't find any considerable advantage of using this model in practice, in fact the performance decreased in many cases. Introduction of the multiple separate \textit{primary attention} parameters essentially increases the number of parameters to learn and results in increased training time \& slower convergence. This model is also a lesser powerful multi-task model compared to  the original proposed model, as only the input feature set is shared here but the parameters in the \textit{primary attention} module \& the subsequent modules are not shared.
For the proposed \textit{attentive multi-task transfer learning} model, we also experiment with another attention based architecture where the \textit{primary attention} module is applied over the pre-trained models and the \textit{secondary attention} module is applied over the sub-images. Performance of this model is poor compared to the original proposed model (Results are not reported due to space constraints).

\section{Conclusion}
In this work we proposed an attention based multi-task transfer learning model for multi-dimensional visual affect prediction in digital photos. We first extracted features from the images using transfer learning with several pre-trained ImageNet models. We then find the important image-parts \& pre-trained features by applying an attention based module. Finally the multi-dimensional affects were predicted using the multi-task learning framework. We performed a detailed analysis of the proposed model and with our experiments show the importance of each of the components in the overall framework. Our results show large improvements over current state-of-the-art models in the benchmark dataset.

{\small
\bibliographystyle{ieee}
\bibliography{egbib}
}

\end{document}

%% file: main_results.tex
\begin{table*}[h!]
\begin{center}
\caption{Experimental five-fold cross validation results for the proposed \textit{attentive multi-task transfer learning} model when trained against the mean value \& the median value of the annotated labels.}
\label{result-multi-mean}
\resizebox{0.98\textwidth}{!}
{
\begin{tabular}{l||cC{5em}c||cC{5em}c}
\hline \hline
\multirow{2}{*}{\bf Affective Dimension} & \multicolumn{3}{C{17em}||}{\bf Attentive Multi-task Transfer Learning (Mean Labels)} & \multicolumn{3}{|C{17em}}{\bf Attentive Multi-task Transfer Learning (Median Labels)} \\ \cline{2-7}
& {\bf R-Squared} & {\bf Pearson Correlation} & {\bf RMSE} &{\bf R-Squared} & {\bf Pearson Correlation} & {\bf RMSE}\\ \cline{2-7}
 \hline \hline
Complexity & 0.31 &  0.56 & 0.09 & 0.19 & 0.44 & 0.10\\  
Quality & 0.51 &  0.71 &  0.08 & 0.37 & 0.60 & 0.10\\ 
Appeal & 0.43 &  0.67 &  0.11 & 0.44 & 0.66 & 0.12\\
Naturalness & 0.52 &  0.72 & 0.12 & 0.50 & 0.70 & 0.16\\ 
Pleasantness & 0.55 &  0.74 &  0.10 & 0.37 & 0.61 & 0.12\\ 
Arousal & 0.40 &  0.64 &  0.08 & 0.43 & 0.65 & 0.10\\
Familiarity & 0.27 &  0.52 &  0.12 & 0.19 & 0.43 & 0.16\\
Coping Potential & 0.31 &  0.56 &  0.10 & 0.21 & 0.45 & 0.13\\
Comprehensibility & 0.26 &  0.50 &  0.10 & 0.23 & 0.49 & 0.12\\
Coherence & 0.26 & 0.51 & 0.09 & 0.27 & 0.53 & 0.10\\ 
Excitingness & 0.33 &  0.57 &  0.11 & 0.37 & 0.60 & 0.10\\
General Interest & 0.41 &  0.63 & 0.11 & 0.38 & 0.61 & 0.14\\
\hline\hline
\end{tabular}
}
\end{center}
\end{table*}

\begin{table*}[h!]
\begin{center}
\caption{Experimental five-fold cross validation results for the baseline \textit{attentive task-specific transfer learning} models \& \textit{non-attentive multi-task transfer learning} model when trained against the mean value of the annotated labels.}
\label{result-abla1}
\resizebox{0.98\textwidth}{!}
{
\begin{tabular}{l||cC{5em}c||cC{5em}c}
\hline \hline
\multirow{2}{*}{\bf Affective Dimension} & \multicolumn{3}{C{17em}||}{\bf Attentive Task-specific Transfer Learning (Mean Labels)} & \multicolumn{3}{|C{17em}}{\bf Non-attentive Multi-task Transfer Learning (Mean Labels)} \\ \cline{2-7}
& {\bf R-Squared} & {\bf Pearson Correlation} & {\bf RMSE} &{\bf R-Squared} & {\bf Pearson Correlation} & {\bf RMSE}\\ \cline{2-7}
 \hline \hline
Complexity & 0.26 & 0.51 & 0.12 & 0.16 & 0.40 & 0.15\\  
Quality & 0.47 &  0.68 &  0.11 & 0.31 & 0.56 & 0.16\\ 
Appeal & 0.39 &  0.62 &  0.12 & 0.26 & 0.50 & 0.15\\
Naturalness & 0.49 & 0.69 & 0.14 & 0.36 & 0.59 & 0.16\\ 
Pleasantness & 0.48 &  0.69 &  0.12 & 0.37 & 0.61 & 0.15\\ 
Arousal & 0.31 & 0.57 &  0.11 & 0.27 & 0.52 & 0.14\\
Familiarity & 0.22 &  0.47 &  0.12 & 0.16 & 0.41 & 0.17\\
Coping Potential & 0.26 & 0.50 & 0.14 & 0.17 & 0.42 & 0.15\\
Comprehensibility & 0.22 &  0.46 &  0.11 & 0.16 & 0.39 & 0.13\\
Coherence & 0.23 & 0.47 & 0.10 & 0.11 & 0.32 & 0.14\\ 
Excitingness & 0.28 & 0.52 &  0.13 & 0.22 & 0.47 & 0.14\\
General Interest & 0.36 & 0.60 & 0.12 & 0.25 & 0.50 & 0.16\\
\hline\hline
\end{tabular}
}
\end{center}
\end{table*}

\begin{table*}[h!]
\begin{center}
\caption{Experimental five-fold cross validation results for the baseline \textit{attentive multi-task non-transfer learning} model \& \textit{non-attentive task-specific non-transfer learning} models when trained against the mean value of the annotated labels.}
\label{result-abla2}
\resizebox{0.95\textwidth}{!}
{
\begin{tabular}{l||cC{5em}c||cC{5em}c}
\hline \hline
\multirow{2}{*}{\bf Affective Dimension} & \multicolumn{3}{C{17em}||}{\bf Attentive Multi-task Non-transfer Learning (Mean Labels)} & \multicolumn{3}{|C{17em}}{\bf Non-attentive Task-specific Non-transfer Learning (Mean Labels)} \\ \cline{2-7}
& {\bf R-Squared} & {\bf Pearson Correlation} & {\bf RMSE} &{\bf R-Squared} & {\bf Pearson Correlation} & {\bf RMSE}\\ \cline{2-7}
 \hline \hline
Complexity & 0.15 &  0.39 & 0.14 & 0.08 & 0.29 & 0.17\\  
Quality & 0.30 &  0.54 &  0.14 & 0.20 & 0.45 & 0.18\\ 
Appeal & 0.28 &  0.53 &  0.15 & 0.17 & 0.41 & 0.18\\
Naturalness & 0.37 & 0.60 & 0.16 & 0.27 & 0.52 & 0.19\\ 
Pleasantness & 0.59 &  0.62 &  0.14 & 0.31 & 0.56 & 0.18\\ 
Arousal & 0.25 &  0.51 &  0.15 & 0.20 & 0.45 & 0.19\\
Familiarity & 0.16 &  0.39 &  0.15 & 0.10 & 0.30 & 0.20\\
Coping Potential & 0.17 & 0.40 & 0.17 & 0.11 & 0.31 & 0.18\\
Comprehensibility & 0.17 &  0.39 &  0.13 & 0.13 & 0.34 & 0.17\\
Coherence & 0.14 & 0.37 & 0.12 & 0.10 & 0.32 & 0.17\\ 
Excitingness & 0.21 & 0.45 &  0.15 & 0.15 & 0.38 & 0.18\\
General Interest & 0.22 & 0.47 & 0.16 & 0.21 & 0.46 & 0.19\\
\hline\hline
\end{tabular}
}
\end{center}
\end{table*}

%% file: comparative_results.tex
\begin{table*}[h!]
\begin{center}
\caption{Comparison of proposed \textit{attentive multi-task transfer learning} 
model with the state-of-the-art model in Soleymani. \cite{soleymani2015quest}. The models are trained against the mean value of the annotated labels. Our models performs better than Soleymani. \cite{soleymani2015quest} for all the affective dimensions across all the performance metrics (R-Squared and Pearson Correlation - the larger the better; RMSE - the smaller the better).}
\label{result-compare}
\resizebox{0.95\textwidth}{!}
{
\begin{tabular}{l||cC{5em}c||cC{5em}c}
\hline \hline
\multirow{2}{*}{\bf Affective Dimension} & \multicolumn{3}{C{17em}||}{\bf Attentive Multi-task Transfer Learning (Mean Labels)} & \multicolumn{3}{|C{17em}}{\bf Soleymani. \cite{soleymani2015quest} (Mean Labels)} \\ \cline{2-7}
& {\bf R-Squared} & {\bf Pearson Correlation} & {\bf RMSE} &{\bf R-Squared} & {\bf Pearson Correlation} & {\bf RMSE}\\ \cline{2-7}
 \hline \hline
Complexity & \bf 0.31 & \bf 0.56 & \bf 0.09 & 0.13 & 0.37 & 0.12\\  
Quality & \bf 0.51 & \bf 0.71 & \bf 0.08 & 0.26 & 0.51 & 0.13\\ 
Naturalness & \bf 0.52 & \bf 0.72 & \bf 0.12 & 0.32 & 0.57 & 0.18\\ 
Pleasantness & \bf 0.55 & \bf 0.74 & \bf 0.10 & 0.16 & 0.40 & 0.15\\ 
Arousal & \bf 0.40 & \bf 0.64 & \bf 0.08 & 0.37 & 0.14 & 0.09\\
Familiarity & \bf 0.27 & \bf 0.52 & \bf 0.12 & 0.08 & 0.30 & 0.14\\
Coping Potential & \bf 0.31 & \bf 0.56 & \bf 0.10 & 0.06 & 0.27 & 0.11\\
General Interest & \bf 0.41 & \bf 0.63 & \bf 0.11 & 0.20 & 0.44 & 0.13\\
\hline\hline
\end{tabular}
}
\end{center}
\end{table*}